# Automated binary classification of hazelnut X-ray images: A deep-learning benchmark for quality assessment

G. Sportelli[1,2†], N. Belcari[1,2†], R. Pace[3*], U. Bernardo[4], S. Sultana[1], A. Toncelli[1], M. Giaccone[3]

† These authors contributed equally to this work.
* Corresponding author: robertapace@cnr.it.

[1] Department of Physics "E. Fermi", University of Pisa, Largo Bruno Pontecorvo 3, 56127 Pisa, Italy
[2] National Institute for Nuclear Physics (INFN), Pisa Division, Largo Bruno Pontecorvo 3, 56127 Pisa, Italy
[3] Institute for Agriculture and Forestry Systems in the Mediterranean, National Research Council of Italy (ISAFOM-CNR), Piazzale Enrico Fermi 1, Località Porto del Granatello, 80055 Portici, Naples, Italy
[4] Institute for Sustainable Plant Protection, National Research Council of Italy (IPSP-CNR), Piazzale Enrico Fermi 1, 80055 Portici, Naples, Italy

# Abstract

Non-destructive X-ray imaging can reveal internal hazelnut defects that are difficult to detect by external inspection alone; however, automated interpretation remains challenging because of subtle radiographic differences among classes, marked class imbalance, and limited annotated data. Here, we present a benchmark for binary hazelnut quality classification (healthy versus defective) based on 799 segmented single-kernel X-ray images (224 × 224 pixels, grayscale), grouped into 101 acquisition units. Seven single-model configurations and ten probability-aggregation ensembles were evaluated using a group-wise split-rotation protocol across five data splits generated using different random seeds. Decision thresholds were selected on the validation set, and performance was assessed deterministically on validation and test sets. Under the expert-reassessed annotation condition, the average-probability ensemble of the binary cross-entropy-trained convolutional neural network and frozen Swin Transformer achieved the highest mean balanced accuracy (86.3% ± 1.8%, five seeds), with several other ensembles providing comparable performance. Across methods, substantial split-to-split variability was observed, indicating that multi-split evaluation is essential for reliable model comparison at this dataset scale. Expert reassessment of 15 kernels changed mean balanced accuracy by up to 3.4 percentage points (mean +1.4 pp across 17 methods), while having no material effect on cross-split variance. The results highlight both the potential of deep learning for automated X-ray-based hazelnut quality assessment and the importance of rigorous evaluation and label curation in small, imbalanced agricultural imaging datasets.

# 1. Introduction

Hazelnut (*Corylus avellana* L.) is a high-value tree nut crop whose commercial quality depends strongly on kernel integrity and on the absence of internal defects (Pedrotti et al., 2021; Giaccone et al., 2026). This requirement is particularly relevant within the Pontic–Caucasian hazelnut production system, where Türkiye is the world's leading producer and, together with Georgia and Azerbaijan, accounts for approximately 69% of global hazelnut production (FAOSTAT, 2026). In this region, *Corylus avellana* var. pontica, across its various cultivars, is the most widely represented taxon (Mirotadze et al., 2009). Among these cultivars, "Anakliuri" is of particular industrial interest and therefore represents a relevant case study for the development of automated quality assessment methods. Compared with *Corylus avellana* var. *avellana*, *C. avellana* var. *pontica* is characterized by longer husks that more completely enclose the nut, a trait that can complicate post-harvest handling.

Among the defects of greatest industrial relevance, stink bug damage and rotten kernels are particularly problematic because they directly affect marketability and processing suitability and are among the main determinants of lot downgrading and reduced economic value (Valeriano et al., 2022; Spataro et al., 2024). For example, in the Italian market, 2025 purchase quotations for in-shell hazelnuts indicate that increasing the incidence of hidden rot from ≤ 1% to ≤ 5% is associated with a price reduction of approximately 27%, confirming the substantial economic impact of rotten kernels (Bio Nocciola, 2026).

In the Pontic–Caucasian production context, delays in drying and processing may further promote quality deterioration, particularly the development of rotten defects associated with fungal infection and the potential risk of mycotoxin contamination, including aflatoxins (Battilani et al., 2018; Camardo Leggieri et al., 2022; Salvatore et al., 2023; Aghayev et al., 2025). This makes rapid and reliable internal quality assessment especially relevant not only for commercial grading but also for food safety management. Stink bug feeding represents another major source of kernel deterioration in hazelnut-producing areas and may also interact with fungal spoilage, as feeding injuries can facilitate secondary fungal colonization of damaged kernels (Scarpari et al., 2018).

Previous studies have demonstrated that commercial hazelnuts are susceptible to stink bug damage throughout kernel development (Hedstrom et al., 2014; de Benedetta et al., 2023). Beyond visible deterioration, stink bug attack can also result in hidden internal defects and broader metabolomic alterations that may not be detectable by external inspection alone, further complicating quality assessment (Ozdemir et al., 2023; Squara et al., 2024; de Benedetta et al., 2026).

Because many relevant defects are internal, conventional quality assessment based on kernel inspection remains informative, but it is labour-intensive, time-consuming, and poorly suited to inline industrial sorting (Pannico et al., 2015). For this reason, non-destructive imaging methods have attracted increasing interest in fruit and nut quality control. In particular, X-ray radiography is well suited to the inspection of dense biological structures because it can reveal internal organization, tissue alterations, and other structural anomalies while remaining compatible with rapid industrial workflows (Kotwaliwale et al., 2014; Mahanti et al., 2022; Brandoli et al., 2025).

Deep learning has been increasingly applied to food classification, grading, and safety assessment, although dataset quality, model reliability, and real-time deployment remain important challenges (Zhao et al., 2025). Recent studies have combined X-ray radiography with deep learning or anomaly-detection methods to identify internal disorders in pears and apples (Van De Looverbosch et al., 2022; Tempelaere et al., 2023, 2024), internal rot in avocados (Matsui et al., 2023), and internal defects in walnuts, seeds, camellia seeds, and chestnuts (Zhang et al., 2023a; Zhang et al., 2024; Hamdy et al., 2024; Chen et al., 2025; Ma et al., 2025; Ma et al., 2026). Together, these studies demonstrate the broader potential of X-ray-based artificial intelligence for automated internal quality assessment across fruit, nut, and seed products (Zhang et al., 2023b). Specifically in hazelnuts, terahertz and X-ray-based approaches have highlighted the need for objective, non-destructive methods capable of detecting hidden defects beyond conventional grading (Gennari et al., 2023).

Publicly available X-ray datasets specifically designed for hazelnut quality classification remain very limited. Early X-ray-based hazelnut studies focused mainly on features derived from grey-level distributions and statistical image descriptors rather than end-to-end deep learning (Colangeli et al., 2014; Khosa and Pasero, 2014). To the best of our knowledge, the dataset released by Mele et al. (2025) in association with the study by Vitale et al. (2025) is currently the only publicly available X-ray dataset specifically designed for deep-learning-based hazelnut quality classification. The study confirmed the potential of pretrained convolutional neural networks for detecting stink bug damage, while also highlighting the difficulty of developing reliable industrial classifiers from a limited number of biological specimens. More generally, these applications frequently involve small sample sizes, class imbalance, and label noise, which may reduce the learnability of subtle defect classes and produce overoptimistic or unstable performance estimates if validation is not carefully designed (Karimi et al., 2020; Shi et al., 2024). These limitations motivate the use of transfer learning, anomaly detection, data augmentation or synthetic data generation, careful label curation, and robust validation strategies (Van De Looverbosch et al., 2022; Tempelaere et al., 2023; Hamdy et al., 2024; Zhao et al., 2025). Deep learning architectures are therefore especially attractive because they can learn complex textural and structural patterns directly from images and have already shown promising performance in food inspection and X-ray-based classification tasks (Ünal and Aktaş, 2023).
Transfer learning from ImageNet-pretrained models is often adopted when domain-specific datasets are limited (Shin et al., 2016; Tajbakhsh et al., 2016; Kim et al., 2022). However, radiographic images differ substantially from natural colour images because they encode density-dependent attenuation rather than surface-reflectance information (van Dael et al., 2019). Moreover, its effectiveness depends on the degree of correspondence between the source and target domains (Zhao et al., 2024). Pretrained representations may not always capture the most informative features for subtle X-ray defect discrimination, especially when fine-tuning data are limited (Mei et al., 2022; Hosseinzadeh Taher et al., 2025). This further reinforces the importance of rigorous evaluation strategies when developing automated classifiers for industrial use (Varoquaux and Cheplygina, 2022; Bradshaw et al., 2023). Within this framework, careful label curation becomes a central methodological requirement rather than a secondary annotation step.
In this study, we apply deep learning models to X-ray images of individual kernels of *C. avellana* var. *pontica*, cv. Anakliuri, to assess the feasibility of robust binary classification between healthy and defective kernels under industrially relevant conditions. We further examine the extent to which classification performance is influenced by dataset characteristics, label quality, and model architecture. The findings may inform the design of reliable and scalable AI-assisted sorting solutions for the hazelnut industry.

# 2. Materials and Methods

## 2.1 Hazelnut samples, quality assessment, and annotation conditions

The dataset consisted of 799 hazelnut kernels (*C. avellana* var. *pontica*) of the Georgian cultivar Anakliuri. The samples were harvested in 2024 from an organic orchard located in Zugdidi, in the Samegrelo region of western Georgia (42°30′29″N, 41°52′21″E), and managed according to the local agronomic practices. The hazelnuts were sampled according to Gennari et al. (2023) from a commercial lot of approximately two tons that was expected to contain a relatively high proportion of defective kernels. After harvest, the hazelnuts were dried to a final moisture content of 6%, determined according to Method 2 of the UNECE standard (UNECE, 2025). From shelling until X-ray acquisition, the kernels were stored under vacuum at a constant temperature of 4 °C to preserve quality and prevent further alterations.
After mechanical shelling and before any destructive opening or cutting, the intact kernels were subjected to X-ray acquisition. Kernel quality was initially assessed by trained personnel through external visual inspection, following the inspection procedure and commercial evaluation criteria described in the UNECE standard for hazelnut kernels (UNECE, 2025). Kernels were assigned to the following externally assessable quality classes: healthy (H), stink bug-damaged (“cimiciato”, C), rotten (R), and oily-appearing alteration (“oil-rancidity”, O). After X-ray imaging, only kernels appearing externally healthy were manually opened and sectioned into four parts to identify internal alterations that could not be detected externally. Kernels showing internal rot in the absence of externally visible symptoms were classified as hidden rotten (HR).
To assess the effect of annotation quality on model performance, two annotation conditions were considered. The initial condition corresponded to the labels assigned during the combined external and internal assessment. Under the initial annotation condition, the dataset contained 138 healthy (17.3%) and 661 defective (82.7%) kernels. Following the first round of model training and evaluation, a subset of kernels that had appeared externally healthy but had subsequently been assigned to a defect class after internal inspection underwent targeted expert reassessment. This review revealed that some kernels previously labelled as defective did not exhibit the defects indicated by their original labels. Accordingly, under the reassessed annotation condition, the binary labels of 15 kernels were changed from defective to healthy, while all remaining labels were left unchanged. The 15 kernels relabelled as healthy had originally comprised 7 hidden rotten, 4 stink bug-damaged, and 4 oil-rancidity kernels. Under the reassessed annotation condition, the dataset contained 153 healthy (19.1%) and 646 defective (80.9%) kernels (Table 1). No sample was added, removed, or relabelled outside these 15 reclassifications. The complete training and evaluation protocol was then repeated using the same data partitions, thereby isolating the effect of label reassessment on model performance.

*Table 1. Distribution of hazelnut kernels among the five quality classes (n = 799) under the reassessed annotation condition.*

| Class | Count | Percentage % |
|---|---|---|
| Stink bug-damaged (C) | 424 | 53.1 |
| Healthy (H) | 153 | 19.1 |
| Rotten (R) | 125 | 15.6 |
| Hidden rotten (HR) | 70 | 8.8 |
| Oil-rancidity (O) | 27 | 3.4 |
| **Total** | **799** | **100.0** |

## 2.2 X-ray imaging specifics

All hazelnut kernels were imaged using the Computed Tomography (CT) component of a preclinical SPECT/CT scanner (MILabs U-SPECT/CT). The CT subsystem (MILabs U-CT) comprises a microfocus X-ray source operating in the 10–65 kV range and a high-resolution gadolinium-oxysulfide / CMOS (GOS/CMOS) flat-panel detector with a native 1536 × 1944 pixel array. A single planar projection was performed using a source-to-detector distance (SDD) of 298.04 mm and a source-to-isocentre distance (SID) of 222.42 mm corresponding to a geometric magnification (SDD / SID) of 1.34. The tube voltage was set to 65 kV with a current of 0.25 mA and an exposure of 35 ms. The beam was filtered with 500 µm of Al. A 2 × 2 binning was applied to all images, resulting in 768 × 972-pixel images with a 16-bit depth.

## 2.3 Dataset and task definition

The dataset consisted of the 799 segmented X-ray images of individual hazelnut kernels grouped into 101 acquisition units, each acquisition unit corresponding to one original X-ray scan containing up to eight kernels (Figures 1 and 2); the corresponding identifier in the released dataset is the sample_id column. Images were stored as 224 × 224 pixel grayscale PNG files.

Grouping was required because kernels belonging to the same acquisition unit shared imaging conditions and could exhibit correlated appearance.

For the primary task, the label was mapped to a binary target, with healthy kernels as the negative class and all defect classes as the positive class (Figure 3). In the binary encoding, y = 1 denoted defective kernels whereas y = 0 denoted healthy kernels.

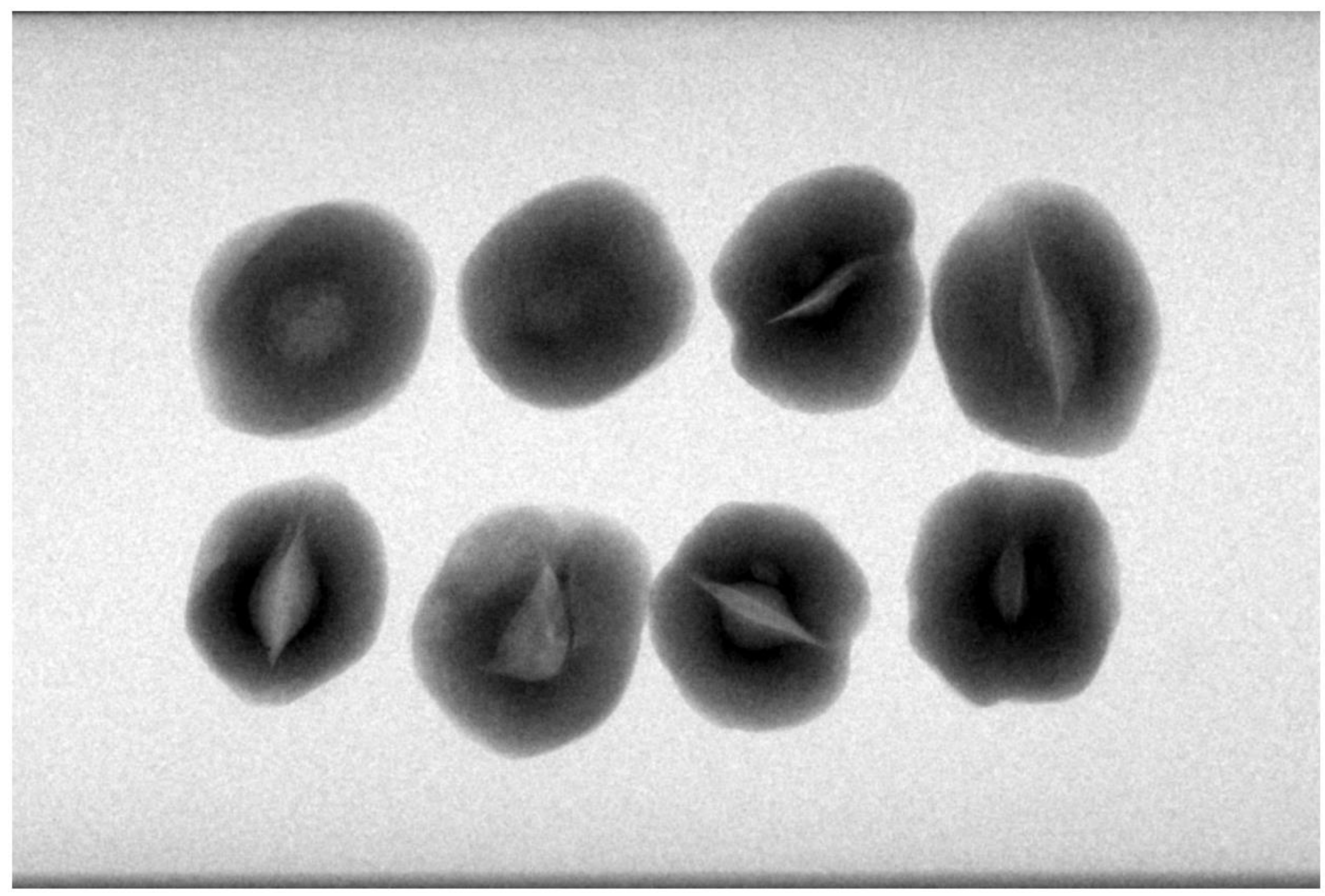

*Figure 1. CT preview image from the X-ray scanner showing eight hazelnuts in a single acquisition before segmentation.*

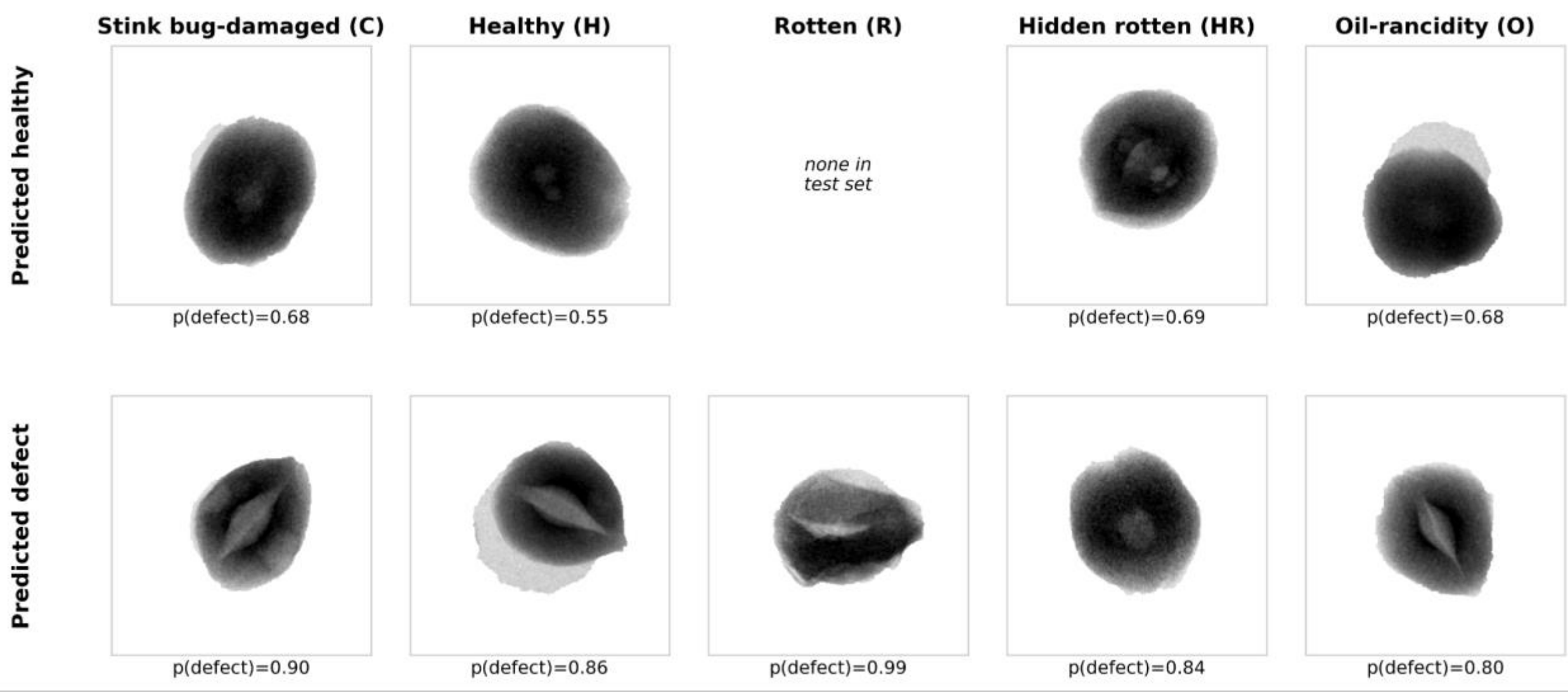

*Figure 2. Test-set examples arranged according to the true quality class (columns, as defined in Table 1) and binary model prediction (rows). Predictions were obtained from the top-ranked ensemble (average-probability aggregation of BinaryNutCNN with binary cross-entropy loss and frozen Swin Transformer Tiny) on split seed 42. Each image is annotated with the ensemble's estimated probability of defect. Grey cells indicate class/prediction combinations absent from the test fold.*

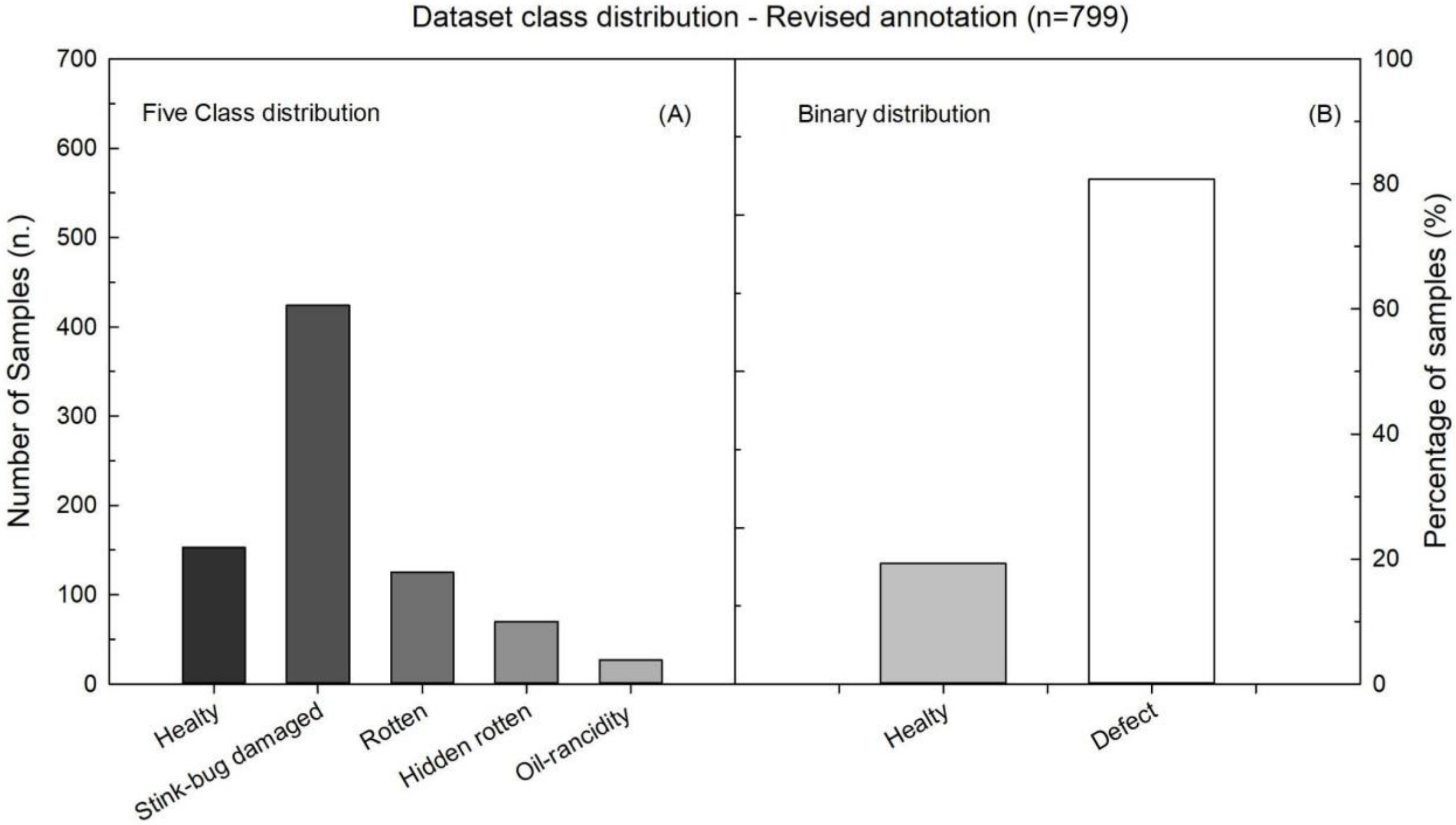

*Figure 3. Distribution of samples across the five quality classes and the corresponding binary grouping (healthy versus defective) used for the benchmark.*

## 2.4 Preprocessing, group-wise splitting, and augmentation

All images are resized to 224 × 224 px and normalized. For models requiring three-channel input, the grayscale image is replicated across channels prior to applying standard ImageNet normalization. Augmentation is limited to random horizontal and vertical flips applied only to the training set, reflecting the arbitrary orientation of segmented nuts while avoiding transformations that may distort subtle radiographic cues. Geometric transformations such as affine warping are excluded because they can alter the apparent thickness profile and edge geometry that encode density-dependent attenuation in X-ray images, potentially introducing artifacts unrelated to genuine anatomical variation.

To prevent data leakage, splitting is performed at the sample_id (group) level with a 70%/15%/15% train/validation/test partition. Stratification is applied based on the majority class within each sample_id group. To ensure that both annotation conditions are evaluated on identical data partitions, stratified splitting is always performed using the initial-condition labels; the reassessed condition inherits these splits without regeneration. This prevents changes in per-group majority labels from altering the fold assignments between conditions, which would confound the comparison. This procedure is referred to as split rotation: the dataset is partitioned multiple times using different random seeds, and each model configuration is trained and evaluated independently on every partition, so that performance statistics reflect variability due to the data split rather than a single arbitrary partition. The model initialization seed is held constant across runs in order to isolate variability attributable to the data partition. Validation and test evaluation are deterministic: no stochastic augmentation is applied and no sampling-based rebalancing is used for validation or test loaders.

As a supplementary sensitivity analysis, the complete benchmark was repeated using nut-level splitting, treating each of the 799 images as an independent sample rather than grouping by acquisition unit. This configuration allowed kernels from the same acquisition unit to be distributed across the training, validation, and test subsets. The group-wise protocol was retained as the primary methodology.

## 2.5 Models and ensembles

Seven single-model configurations are evaluated. These include BinaryNutCNN, a lightweight custom convolutional neural network (CNN) with approximately 0.42 million parameters, and three ImageNet-pretrained backbones adapted for binary classification: Swin Transformer Tiny, EfficientNet-B0, and ResNet-18. BinaryNutCNN is trained from scratch on grayscale input. For the Swin Transformer, both full fine-tuning and a frozen-backbone variant are included to assess the impact of domain mismatch and model capacity.

BinaryNutCNN consists of four convolutional blocks, each comprising a 3×3 convolution, batch normalization, and ReLU activation, with channel widths of 32, 64, 128, and 256; the first three blocks apply 2×2 max-pooling for spatial reduction, and the fourth applies global average pooling. The classifier head is a two-layer fully connected network (256→128→1) with ReLU activation and 50% dropout, producing a single logit for binary classification. Three loss configurations are evaluated for the custom CNN: (i) standard binary cross-entropy (BCE), (ii) focal loss (alpha = 0.75, gamma = 1.0), which down-weights well-classified examples to focus training on hard cases, and (iii) BCE with pos_weight = 2.0, which doubles the loss contribution of positive (defect) samples to increase sensitivity to false negatives. Note that the pos_weight = 2.0 configuration deliberately stacks loss-level and sampling-level rebalancing: as described in Section 2.6, all training runs use a WeightedRandomSampler that already draws healthy and defect samples with equal expected frequency.

Ten ensembles are constructed by combining a custom-CNN configuration with a pretrained backbone using probability aggregation at inference. Specifically, the following five pairs are considered: (i) BinaryNutCNN (BCE) with Swin Transformer Tiny (full); (ii) BinaryNutCNN (BCE) with Swin Transformer Tiny (frozen); (iii) BinaryNutCNN (BCE) with EfficientNet-B0; (iv) BinaryNutCNN (focal) with Swin Transformer Tiny (full); (v) BinaryNutCNN (focal) with Swin Transformer Tiny (frozen). Each pair is aggregated with two rules: average probability and maximum probability, yielding ten ensemble configurations. Aggregation operates on sigmoid probabilities: each constituent model's raw logit output is passed through a sigmoid function, and the resulting probabilities are combined by element-wise averaging or element-wise maximum. The decision threshold is then selected on the aggregated validation-set probabilities using the same grid-search procedure as for single models.

*Table 2. Overview of the evaluated model families and representative capacities.*

| Family | Input | Trainable parameters | Notes |
|---|---|---|---|
| BinaryNutCNN (BCE) | Gray | ~ 0.42M | Trained from scratch; standard BCE loss |
| BinaryNutCNN (focal) | Gray | ~ 0.42M | Trained from scratch; focal loss ($\alpha$=0.75, $\gamma$=1.0) |
| BinaryNutCNN (pw2) | Gray | ~ 0.42M | Trained from scratch; BCE with pos_weight=2.0 |
| Swin Transformer Tiny (full) | RGB | ~ 27.5M | End-to-end fine-tuning |
| Swin Transformer Tiny (frozen) | RGB | ~ 770 | Frozen backbone; only the linear classification head is trained |
| EfficientNet-B0 (full) | RGB | ~ 4.0M | End-to-end fine-tuning |
| ResNet-18 (full) | RGB | ~ 11.2M | End-to-end fine-tuning |

## 2.6 Training and evaluation protocol

Models are trained for up to 50 epochs with early stopping (patience 15 epochs, monitoring validation balanced accuracy). Class imbalance during training is addressed by a WeightedRandomSampler that assigns each sample a weight equal to the inverse of its class count in the training fold, so that healthy and defect samples are drawn with equal expected frequency; sampling is with replacement, with the number of draws per epoch equal to the training set size.
CNN models use AdamW with learning rate $3\times10^{-4}$ and weight decay $10^{-4}$, with a ReduceLROnPlateau scheduler (factor 0.5, patience 5 epochs). Pretrained backbones (EfficientNet-B0, ResNet-18) use AdamW with learning rate $10^{-4}$ and weight decay $10^{-4}$, also with ReduceLROnPlateau. The Swin Transformer configurations use AdamW with learning rate $10^{-4}$ (full fine-tuning) or $10^{-3}$ (frozen backbone), weight decay 0.01, BCEWithLogitsLoss with pos_weight = 3.0, and a cosine-annealing schedule with linear warmup over 5 epochs. Gradient clipping is applied to all models with max norm 2.0.

*Table 3. Key training and evaluation settings shared across methods.*

| Item | Setting |
|---|---|
| Data splits | Group-wise by sample_id, 70%/15%/15% train/validation/test, stratified by majority label per group |
| Split rotation | Five split seeds with a fixed model-initialization seed |
| Augmentation | Random horizontal and vertical flips applied to the training set only |
| Early stopping | Patience 15 epochs based on validation balanced accuracy |
| Training | Up to 50 epochs with early stopping |
| Threshold selection | Validation-set grid search from 0.10 to 0.90 in steps of 0.02 |
| Primary metric | Balanced accuracy |
| Secondary metrics | Healthy recall, defect recall, false positives, and false negatives |

## 2.7 Hardware

All experiments were run on a Linux workstation (Ubuntu 20.04.6 LTS, kernel 5.4) equipped with an AMD EPYC 7543 32-core CPU (64 threads), 1 TiB of system RAM, and four NVIDIA A100-SXM4 accelerators with 40 GB of on-board memory each (NVIDIA driver 575.57.08, CUDA runtime 12.9). Training and evaluation were implemented in Python 3.12 with PyTorch 2.10 (CUDA 12.8 build), torchvision 0.25 and cuDNN 9.10, and used a single A100 GPU per run (no data or model parallelism). The complete benchmark (17 methods × 5 split seeds × 2 annotation conditions) ran end-to-end in 38.4 minutes of wall-clock time.

# 3. Results

## 3.1 Benchmark performance under the reassessed annotation condition

Under the expert-reassessed annotation condition, ensemble configurations occupied the seven highest-ranking positions. The best-performing method was the average-probability ensemble combining the BCE-trained CNN and the frozen Swin Transformer (ens_avg_bce_frozen), which achieved a mean balanced accuracy of 86.3% ± 1.8% across the five splits (Table 4). This configuration also showed one of the lowest levels of cross-split variability among the evaluated methods. The remaining top-ranked ensembles achieved mean balanced accuracies ranging from 83.1% to 85.1%.

For ens_avg_bce_frozen, mean defect recall was 79.4% ± 2.0%, whereas mean healthy recall was 93.1% ± 5.0%. The corresponding absolute error counts averaged 20.6 ± 2.0 false negatives per split (range: 18–24) and 1.8 ± 1.2 false positives per split (range: 0–3), with test folds comprising 126–128 kernels. Validation-selected decision thresholds ranged from 0.42 to 0.78, indicating that the selected operating point varied across validation-set compositions (Table S4).

*Table 4. Top-ranked methods under the reassessed annotation condition (balanced accuracy, five-seed mean ± standard deviation).*

| Rank | Method | Mean | Std |
|---|---|---|---|
| 1 | ens_avg_bce_frozen | 86.3% | 1.8% |
| 2 | ens_avg_focal_frozen | 85.1% | 2.9% |
| 3 | ens_max_bce_full | 83.8% | 4.0% |
| 4 | ens_avg_effnet | 83.4% | 4.5% |
| 5 | ens_max_bce_frozen | 83.3% | 4.6% |
| 6 | ens_avg_focal_full | 83.1% | 3.6% |
| 7 | ens_max_effnet | 83.1% | 4.5% |

## 3.2 Sensitivity to expert reassessment

Mean balanced accuracy differed between the two annotation conditions by −0.5 to +3.4 percentage points depending on the configuration. Selected configurations are summarized in Table 5; complete per-seed results under the reassessed and initial annotation conditions are reported in Tables S1 and S2 respectively, with the corresponding per-seed differences in Table S3. Thirteen of the seventeen evaluated methods showed a positive mean change (up to +3.4 pp for ens_max_focal_full and +3.1 pp for ens_avg_bce_frozen), while four methods showed small negative changes within one standard deviation of zero (−0.1 to −0.5 pp). At the individual-split level the effect was not uniform: on some seeds the reassessed condition performed slightly worse than the initial one (Table S3). Cross-split variability was essentially unchanged between conditions (average standard deviation 3.44% under the initial condition and 3.46% under the reassessed condition), with 9 methods showing a marginal increase and 8 a marginal decrease.

*Table 5. Selected method performance under two annotation conditions (balanced accuracy, five-seed mean ± standard deviation).*

| Method | Initial condition | Reassessed condition | Delta |
|---|---|---|---|
| swin_full | 80.4% ± 5.2% | 82.6% ± 3.7% | +2.2% |
| ens_avg_bce_full | 82.1% ± 3.7% | 82.9% ± 1.6% | +0.8% |
| cnn_bce | 79.7% ± 4.4% | 81.5% ± 3.4% | +1.8% |
| ens_avg_effnet | 83.7% ± 1.4% | 83.4% ± 4.5% | −0.3% |
| Efficientnet | 81.5% ± 3.1% | 82.8% ± 3.4% | +1.4% |
| swin_frozen | 81.3% ± 2.5% | 81.2% ± 2.2% | −0.1% |
| ens_avg_bce_frozen | 83.1% ± 3.4% | 86.3% ± 1.8% | +3.1% |

## 3.3 Split-to-split variability

Balanced accuracy varied systematically across data splits under the expert-reassessed annotation condition (Figure 4). For the top-ranked method, ens_avg_bce_frozen, performance ranged from 83.8% to 88.9%, corresponding to a span of 5.1 percentage points, whereas the other methods shown exhibited wider ranges. Despite differences in their absolute performance, the methods tended to improve or decline on the same splits. This co-variation suggests that, at this dataset scale, evaluation performance is strongly influenced by test-fold composition, particularly by which acquisition units contribute healthy samples, in addition to differences among model architectures.

No method achieved the highest balanced accuracy on every split. Moreover, among the leading configurations, split-to-split variability was of the same order as the small differences in their mean balanced accuracy. Accordingly, the ordering in Table 4 should be interpreted as a ranking based on the five-split means rather than as evidence of consistent per-split dominance (Table S1).

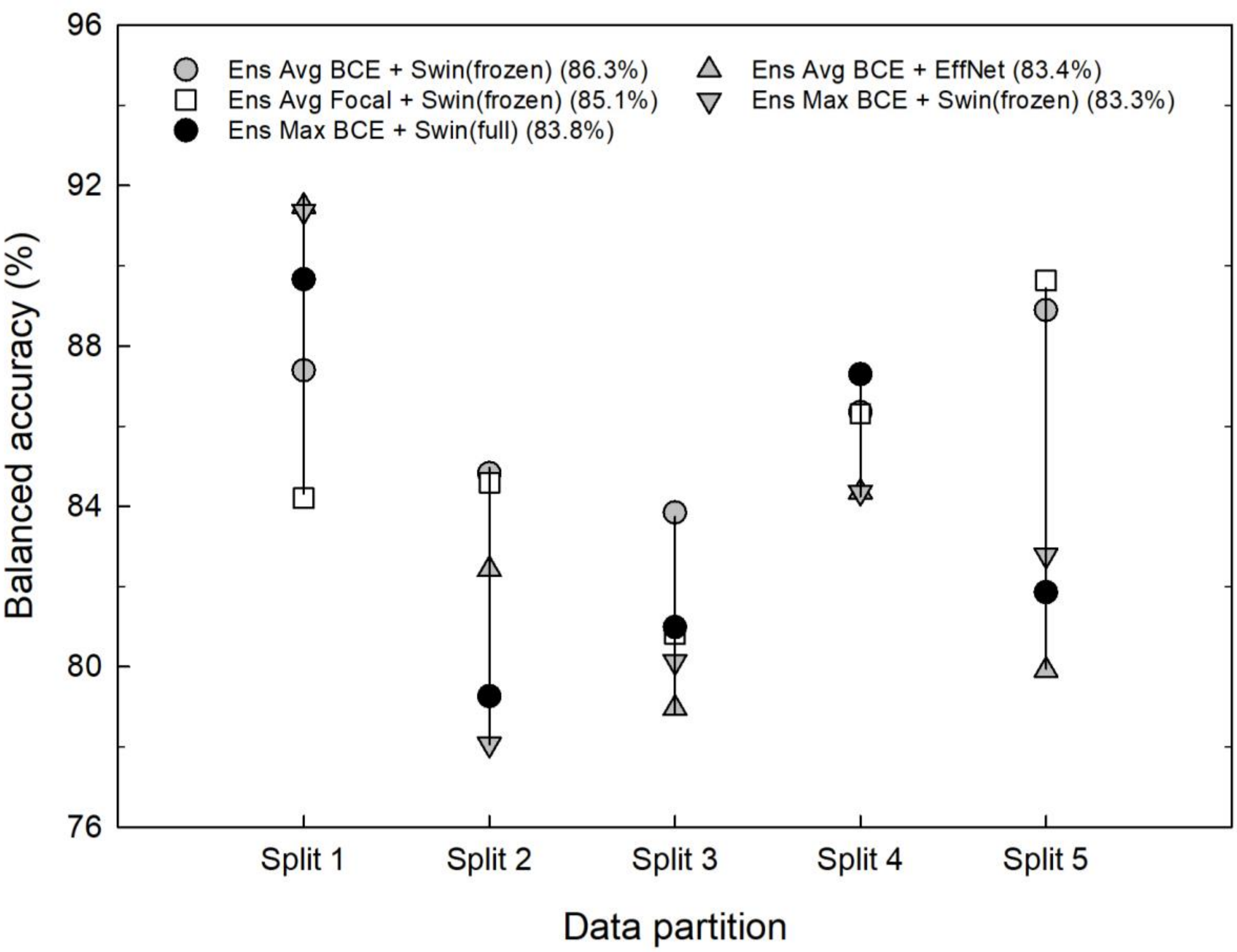

*Figure 4. Per-split balanced accuracy for the five top-ranked methods under the reassessed annotation condition. Each marker represents one method, and each x-axis position represents one data partition; marker shapes identify methods as indicated in the legend.*

## 3.4 Sensitivity to the data-partitioning strategy

As a supplementary sensitivity analysis, the complete benchmark was repeated using nut-level splitting, which allowed kernels originating from the same acquisition unit to be distributed across the training, validation, and test subsets. Across the 17 evaluated methods, mean balanced accuracy was 81.5% under nut-level splitting and 82.8% under group-wise splitting. Thus, sharing acquisition units across data subsets did not yield higher average performance in this analysis, providing no indication that acquisition-specific characteristics produced an optimistic performance advantage. The group-wise protocol was therefore retained as the primary evaluation strategy because it prevents acquisition-level information leakage and provides a more conservative estimate of generalization to previously unseen acquisitions.

# 4. Discussion

## 4.1 Industrial rationale, X-ray ambiguity, and label quality

The healthy-versus-defective formulation adopted in this study reflects the primary operational decision in industrial sorting: whether a kernel can remain in the production stream or should be rejected, downgraded, or redirected for further inspection, rather than whether it can be assigned to a specific pathological or physiological category (Patel et al., 2011; Moscetti et al., 2015). This framing is appropriate for commercial lots in which rotten kernels, hidden rot, stink bug damage, and oil-rancidity-related alterations may coexist and lead to similar consequences for commercial value, processing suitability, and sensory quality (Memoli et al., 2017; Pedrotti et al., 2021; de Benedetta et al., 2023; Spataro et al., 2024; de Benedetta et al., 2026). Moreover, the pronounced imbalance among the original five quality classes, with some defect categories represented by relatively few samples (Figure 3), would make multiclass modelling at the present dataset scale prone to unstable class-specific estimates and overinterpretation of performance (Brishti et al., 2025; Steinert et al., 2024). The binary benchmark therefore provides a more reliable assessment of model feasibility without implying defect-specific diagnosis.

Aggregation into a single defective class does not, however, eliminate the biological and radiographic heterogeneity of the samples. The defects arise from distinct but sometimes interconnected processes: stink bug feeding may alter kernel tissues and facilitate secondary microbial colonization, whereas rot and oil-rancidity-related alterations involve different structural and biochemical changes (Scarpari et al., 2018; Memoli et al., 2017; de Benedetta et al., 2023; de Benedetta et al., 2026). In two-dimensional X-ray images, these processes may produce weak, localized, or partially overlapping patterns, including subtle density variations, small cavities, tissue discontinuities, and changes in internal texture (Gennari et al., 2023). As illustrated in Figure 2, kernels belonging to different defect classes may therefore show similar radiographic appearances, while mildly affected kernels may remain difficult to distinguish from healthy ones even under a binary formulation.

The expert reassessment nevertheless highlighted an important aspect of reference-label construction. Despite the two-stage physical inspection protocol combining external UNECE grading and destructive sectioning of externally healthy kernels, a small subset of internal-defect cases remained ambiguous upon subsequent review. This suggests that targeted expert reassessment can provide an additional quality-control step when developing datasets in which internal defect categories are inherently difficult to define.

## 4.2 Model performance and evaluation robustness

Under the expert-reassessed annotation condition, the best-performing method achieved a mean balanced accuracy of 86.3% ± 1.8% across the five data splits. Ensemble configurations occupied the seven highest-ranking positions, although several obtained comparable results. The benchmark therefore supports the usefulness of probability aggregation at the present dataset scale but does not identify a uniquely superior architecture. Notably, ensembles incorporating the frozen Swin Transformer remained competitive, indicating that full backbone fine-tuning was not required to achieve strong performance. More generally, no systematic relationship emerged between trainable parameter count and sensitivity to expert reassessment, suggesting that model capacity alone did not explain the observed performance differences.

Expert reassessment changed mean balanced accuracy by −0.5 to +3.4 percentage points across the 17 methods, with a mean change of +1.4 percentage points (Table S3). Average cross-split variability was essentially unchanged (3.44% under the initial and 3.46% under the reassessed condition), and label quality therefore affected the overall level of measured performance only modestly without altering its dependence on data partitioning. Even for the top-ranked method, balanced accuracy ranged from 83.8% to 88.9% across splits, and different methods tended to improve or decline on the same partitions. This co-variation indicates that test-fold composition, including the limited number of healthy kernels and independent acquisition units, substantially influenced the resulting estimates. Consequently, small differences among the leading methods should not be overinterpreted, and rankings based on a single train–validation–test partition would be unreliable (Table S1; Allgaier and Pryss, 2024; Bradshaw et al., 2023).

The supplementary nut-level analysis provided no evidence that sharing acquisition units across subsets artificially increased performance: mean balanced accuracy was 81.5% under nut-level splitting and 82.8% under group-wise splitting. Nevertheless, group-wise partitioning remains methodologically preferable because it prevents acquisition-level information leakage and evaluates generalization to previously unseen acquisitions. Taken together, these findings identify curated annotations and repeated group-wise evaluation as essential requirements for obtaining reliable performance estimates from small, imbalanced agricultural imaging datasets.

## 4.3 Operational interpretation of errors and thresholds

False positives and false negatives have different consequences in industrial sorting. A false negative corresponds to a defective kernel classified as healthy and potentially retained in the accepted product stream. This is generally the more critical error when the objective is to minimize residual defectiveness, because it may affect lot quality, commercial value, and sensory properties and, for defects associated with rot or fungal contamination, may also raise food-safety concerns (Salvatore et al., 2023; Gavilán-CuiCui et al., 2025; Casu et al., 2026). Conversely, a false positive results in the rejection or reinspection of a healthy kernel, reducing product recovery and increasing operational costs. The acceptable balance between these errors therefore depends on the intended role of the system: a conservative pre-screening application may prioritize defect recall while tolerating more false positives (Komisarenko and Kull, 2025).

In this benchmark, decision thresholds were selected on the validation set to maximize balanced accuracy, providing a consistent criterion for model comparison under class imbalance (Brodersen et al., 2010). These thresholds should not, however, be interpreted as universal industrial operating points. For the top-ranked method, they ranged from 0.42 to 0.78 across splits (Table S4), indicating sensitivity to validation-set composition. Before deployment, the threshold should therefore be recalibrated using independent, production-representative lots and selected according to the acceptable residual defectiveness, desired product recovery, available reinspection capacity, and specific objectives of the sorting line (Esposito et al., 2021; Romaniello et al., 2024).

### 4.4 Limitations and industrial translation

The present study should be interpreted as a controlled benchmark rather than a deployment-ready sorting solution. The dataset comprised 799 segmented kernel images from 101 acquisition units and represented a single cultivar ('Anakliuri', *Corylus avellana* var. *pontica*) and production context. Its biological variability was therefore limited, particularly for minority categories such as hidden rot and oil-rancidity-related alterations. Broader evaluation will require larger and more balanced datasets encompassing different cultivars, production areas, seasons, post-harvest histories, defect severities, and acquisition conditions, with greater representation of healthy kernels and minority defects.

The reported performance is also specific to the experimental X-ray system used to generate the dataset. This configuration, originally developed for three-dimensional small-animal imaging, provided controlled, high-quality images but was not designed for in-line hazelnut sorting. Because image resolution, contrast, noise, energy spectrum, and projection geometry depend on the acquisition chain, the present results should not be regarded as scanner-independent estimates. Industrial translation will require a dedicated, compact, and high-throughput platform in which tube voltage, tube current, exposure time, conveyor speed, radiation dose, defect detectability, and classification performance are optimized jointly. Dual- or multi-energy configurations may also be investigated if they provide additional sensitivity to subtle internal alterations without compromising throughput.

Finally, prospective validation on independent industrial lots is essential to assess generalization and establish stable operating thresholds. Evaluation should extend beyond balanced accuracy to operational outcomes such as residual defectiveness, product recovery, reinspection workload, lot-grading accuracy, and scanning throughput. Such validation will determine whether the proposed approach can provide measurable benefits under realistic production conditions and clarify its most appropriate role as a decision-support component within an industrial quality control system.

## 5. Conclusions

This study provides evidence that two-dimensional X-ray imaging combined with deep-learning ensembles can support the non-destructive separation of healthy and defective hazelnut kernels. Its main contribution lies not in identifying a universally superior architecture, but in establishing a rigorous evaluation framework in which annotation quality and independence among acquisition groups are integral to model assessment. The proposed benchmark therefore provides a methodological basis for future research and for the development of dedicated X-ray sorting systems. Further studies should determine whether this approach can deliver consistent benefits under production-representative conditions.

## Data availability

The dataset supporting this study, including the segmented X-ray images, sample metadata, expert-reassessed annotations, and data partitions used for model training, validation, and testing, is publicly available on Zenodo at https://doi.org/10.5281/zenodo.21739932.

## Declaration of competing interest

The authors declare that they have no known competing financial interests or personal relationships that could have appeared to influence the work reported in this paper.

## Acknowledgments

The authors gratefully acknowledge Pakka Georgia LLC (Zugdidi, Georgia) for supplying the hazelnut samples used in this study.

## Funding

The research was carried out with the cooperation and contribution of the Hazelnut Company division of Ferrero Group (FERRERO S.R.L.).

## Declaration of generative AI

During the preparation of this work, the authors used Generative AI tools to support language editing and improve the clarity, structure, and consistency of the manuscript. After using these tools, the authors reviewed and edited the content as needed and take full responsibility for the content of the published article.

**Author Contributions:** Conceptualization, M.G., N.B., G.S., A.T.; methodology, M.G., R.P., U.B., N.B., G.S., A.T.; validation, M.G., R.P., G.S.; formal analysis, M.G., R.P., G.S.; investigation, M.G., R.P., N.B., S.S.; data curation, M.G., R.P., G.S., S.S.; software, G.S.; visualization, M.G., R.P., G.S.; writing-original draft preparation, M.G., R.P., G.S.; writing-review and editing, M.G., R.P., U.B., N.B., G.S., A.T.; supervision, M.G., U.B., A.T.; funding acquisition, M.G., U.B., A.T. All authors have read and agreed to the published version of the manuscript.

# Supplementary materials

Companion tables to the main article, reporting the full split-rotation benchmark broken down by individual split seed. Values are balanced accuracy on the test fold, expressed in %, computed with the same training and evaluation protocol described in Section 2.6 of the main article. The five split seeds are [42, 123, 456, 789, 2024] and the model-initialization seed is 42 for all runs. Test-fold sizes are 128, 127, 126, 128, 128 kernels for seeds 42, 123, 456, 789 and 2024 respectively (train ≈ 546, val ≈ 126). Stratification is always performed on the initial-condition labels, so each of the two annotation conditions uses identical data partitions. Rows in Tables S1–S3 are ordered by decreasing mean balanced accuracy under the reassessed annotation condition; the first seven correspond to the methods reported in Table 4 of the main article. Mean and standard deviation are computed with population variance (ddof = 0), matching the convention used in Tables 4 and 5.

Table S1. Balanced accuracy per split seed — reassessed annotation condition.

| # | Method | Seed 42 | Seed 123 | Seed 456 | Seed 789 | Seed 2024 | Mean | Std | Min | Max |
|---|---|---|---|---|---|---|---|---|---|---|
| 1 | ens_avg (cnn_bce + swin_frozen) | 87.39 | 84.82 | 83.84 | 86.35 | 88.89 | 86.26 | 1.80 | 83.84 | 88.89 |
| 2 | ens_avg (cnn_focal + swin_frozen) | 84.20 | 84.59 | 80.81 | 86.31 | 89.63 | 85.11 | 2.88 | 80.81 | 89.63 |
| 3 | ens_max (cnn_bce + swin_full) | 89.66 | 79.26 | 80.98 | 87.29 | 81.85 | 83.81 | 3.97 | 79.26 | 89.66 |
| 4 | ens_avg (cnn_bce + effnet) | 91.48 | 82.42 | 78.96 | 84.35 | 79.91 | 83.42 | 4.45 | 78.96 | 91.48 |
| 5 | ens_max (cnn_bce + swin_frozen) | 91.36 | 78.07 | 80.13 | 84.35 | 82.78 | 83.34 | 4.55 | 78.07 | 91.36 |
| 6 | ens_avg (cnn_focal + swin_full) | 87.39 | 78.78 | 80.47 | 87.29 | 81.67 | 83.12 | 3.56 | 78.78 | 87.39 |
| 7 | ens_max (cnn_bce + effnet) | 91.93 | 81.46 | 79.12 | 81.45 | 81.57 | 83.11 | 4.51 | 79.12 | 91.93 |

| # | Method | Seed 42 | Seed 123 | Seed 456 | Seed 789 | Seed 2024 | Mean | Std | Min | Max |
|---|---|---|---|---|---|---|---|---|---|---|
| 8 | ens_avg (cnn_bce + swin_full) | 83.98 | 81.21 | 81.48 | 85.33 | 82.41 | 82.88 | 1.56 | 81.21 | 85.33 |
| 9 | EfficientNet-B0 | 89.20 | 81.46 | 79.12 | 82.88 | 81.57 | 82.85 | 3.40 | 79.12 | 89.20 |
| 10 | Swin-T (full) | 86.82 | 78.78 | 80.47 | 87.29 | 79.63 | 82.60 | 3.68 | 78.78 | 87.29 |
| 11 | ens_max (cnn_focal + swin_full) | 85.11 | 80.00 | 79.97 | 87.29 | 78.24 | 82.12 | 3.46 | 78.24 | 87.29 |
| 12 | ens_max (cnn_focal + swin_frozen) | 83.07 | 87.25 | 76.43 | 82.88 | 79.81 | 81.89 | 3.61 | 76.43 | 87.25 |
| 13 | BinaryNutCNN (pw2) | 86.25 | 84.11 | 74.75 | 80.43 | 82.87 | 81.68 | 3.95 | 74.75 | 86.25 |
| 14 | BinaryNutCNN (BCE) | 87.95 | 79.52 | 77.95 | 81.41 | 80.83 | 81.53 | 3.43 | 77.95 | 87.95 |
| 15 | BinaryNutCNN (focal) | 82.84 | 88.94 | 74.92 | 83.33 | 77.50 | 81.51 | 4.90 | 74.92 | 88.94 |
| 16 | Swin-T (frozen) | 83.07 | 77.59 | 79.63 | 82.88 | 82.78 | 81.19 | 2.20 | 77.59 | 83.07 |
| 17 | ResNet-18 | 84.89 | 79.03 | 81.99 | 82.96 | 76.76 | 81.12 | 2.89 | 76.76 | 84.89 |

Table S2. Balanced accuracy per split seed — initial annotation condition
Row order follows Table S1 (ranked by mean under the reassessed condition), so per-method comparison across conditions is done by reading the same row in S1 and S2.

| # | Method | Seed 42 | Seed 123 | Seed 456 | Seed 789 | Seed 2024 | Mean | Std | Min | Max |
|---|---|---|---|---|---|---|---|---|---|---|
| 1 | ens_avg (cnn_bce + swin_frozen) | 81.01 | 82.52 | 79.89 | 82.63 | 89.55 | 83.12 | 3.37 | 79.89 | 89.55 |
| 2 | ens_avg (cnn_focal + swin_frozen) | 77.98 | 82.05 | 84.00 | 83.58 | 86.31 | 82.78 | 2.76 | 77.98 | 86.31 |
| 3 | ens_max (cnn_bce + swin_full) | 82.95 | 71.52 | 79.40 | 87.44 | 85.35 | 81.33 | 5.59 | 71.52 | 87.44 |
| 4 | ens_avg (cnn_bce + effnet) | 85.44 | 82.01 | 82.52 | 85.08 | 83.59 | 83.73 | 1.35 | 82.01 | 85.44 |
| 5 | ens_max (cnn_bce + swin_frozen) | 84.41 | 86.30 | 84.49 | 80.75 | 81.62 | 83.51 | 2.04 | 80.75 | 86.30 |
| 6 | ens_avg (cnn_focal + swin_full) | 81.01 | 78.19 | 81.81 | 87.44 | 84.90 | 82.67 | 3.20 | 78.19 | 87.44 |
| 7 | ens_max (cnn_bce + effnet) | 84.89 | 84.37 | 79.86 | 85.55 | 83.13 | 83.56 | 2.01 | 79.86 | 85.55 |
| 8 | ens_avg (cnn_bce + swin_full) | 78.10 | 77.72 | 82.29 | 86.96 | 85.35 | 82.08 | 3.73 | 77.72 | 86.96 |
| 9 | EfficientNet-B0 | 83.50 | 82.01 | 79.86 | 85.55 | 76.57 | 81.50 | 3.09 | 76.57 | 85.55 |
| 10 | Swin-T (full) | 80.52 | 71.52 | 79.40 | 87.44 | 83.08 | 80.39 | 5.23 | 71.52 | 87.44 |
| 11 | ens_max (cnn_focal + swin_full) | 79.07 | 71.52 | 79.40 | 87.91 | 75.61 | 78.70 | 5.41 | 71.52 | 87.91 |
| 12 | ens_max (cnn_focal + swin_frozen) | 77.01 | 83.00 | 84.49 | 84.05 | 76.97 | 81.10 | 3.39 | 76.97 | 84.49 |
| 13 | BinaryNutCNN (pw2) | 80.04 | 83.96 | 73.32 | 78.56 | 78.94 | 78.96 | 3.41 | 73.32 | 83.96 |
| 14 | BinaryNutCNN (BCE) | 81.50 | 83.94 | 72.35 | 77.14 | 83.59 | 79.70 | 4.40 | 72.35 | 83.94 |
| 15 | BinaryNutCNN (focal) | 77.13 | 84.43 | 75.26 | 79.97 | 78.94 | 79.15 | 3.09 | 75.26 | 84.43 |
| 16 | Swin-T (frozen) | 77.01 | 82.52 | 84.49 | 80.75 | 81.67 | 81.29 | 2.47 | 77.01 | 84.49 |
| 17 | ResNet-18 | 81.98 | 85.31 | 80.10 | 74.19 | 76.16 | 79.55 | 3.99 | 74.19 | 85.31 |

Table S3. Per-seed delta (reassessed − initial), in balanced-accuracy percentage points. Positive values indicate that the reassessed condition improved performance on that split. Thirteen of the seventeen methods show a positive mean delta and four show a small negative one; at the individual-split level the sign varies within each row, with more mixed behaviour across seeds compared with the ranking observed on the mean.

| # | Method | Seed 42 | Seed 123 | Seed 456 | Seed 789 | Seed 2024 | Mean Δ | Min Δ | Max Δ |
|---|---|---|---|---|---|---|---|---|---|
| 1 | ens_avg (cnn_bce + swin_frozen) | +6.38 | +2.30 | +3.95 | +3.72 | −0.66 | +3.14 | −0.66 | +6.38 |
| 2 | ens_avg (cnn_focal + swin_frozen) | +6.22 | +2.54 | −3.19 | +2.74 | +3.32 | +2.32 | −3.19 | +6.22 |
| 3 | ens_max (cnn_bce + swin_full) | +6.71 | +7.75 | +1.58 | −0.14 | −3.50 | +2.48 | −3.50 | +7.75 |
| 4 | ens_avg (cnn_bce + effnet) | +6.04 | +0.41 | −3.57 | −0.73 | −3.68 | −0.30 | −3.68 | +6.04 |
| 5 | ens_max (cnn_bce + swin_frozen) | +6.96 | −8.23 | −4.35 | +3.61 | +1.16 | −0.17 | −8.23 | +6.96 |
| 6 | ens_avg (cnn_focal + swin_full) | +6.38 | +0.59 | −1.34 | −0.14 | −3.23 | +0.45 | −3.23 | +6.38 |
| 7 | ens_max (cnn_bce + effnet) | +7.04 | −2.91 | −0.74 | −4.10 | −1.56 | −0.45 | −4.10 | +7.04 |
| 8 | ens_avg (cnn_bce + swin_full) | +5.88 | +3.49 | −0.81 | −1.63 | −2.95 | +0.80 | −2.95 | +5.88 |
| 9 | EfficientNet-B0 | +5.71 | −0.55 | −0.74 | −2.67 | +5.01 | +1.35 | −2.67 | +5.71 |
| 10 | Swin-T (full) | +6.29 | +7.27 | +1.07 | −0.14 | −3.45 | +2.21 | −3.45 | +7.27 |
| 11 | ens_max (cnn_focal + swin_full) | +6.05 | +8.48 | +0.57 | −0.61 | +2.63 | +3.42 | −0.61 | +8.48 |

| 12 | ens_max (cnn_focal + swin_frozen) | +6.06 | +4.25 | −8.06 | −1.17 | +2.85 | +0.79 | −8.06 | +6.06 |
|---|---|---|---|---|---|---|---|---|---|
| 13 | BinaryNutCNN (pw2) | +6.21 | +0.15 | +1.43 | +1.87 | +3.93 | +2.72 | +0.15 | +6.21 |
| 14 | BinaryNutCNN (BCE) | +6.46 | −4.42 | +5.59 | +4.27 | −2.75 | +1.83 | −4.42 | +6.46 |
| 15 | BinaryNutCNN (focal) | +5.71 | +4.51 | −0.35 | +3.36 | −1.44 | +2.36 | −1.44 | +5.71 |
| 16 | Swin-T (frozen) | +6.06 | −4.93 | −4.86 | +2.13 | +1.11 | −0.10 | −4.93 | +6.06 |
| 17 | ResNet-18 | +2.91 | −6.28 | +1.89 | +8.77 | +0.60 | +1.58 | −6.28 | +8.77 |

Table S4. Detailed per-seed metrics for the top-ranked method (ens_avg_bce_frozen, reassessed condition). Balanced accuracy, healthy recall, defect recall, absolute false positives / false negatives, and the validation-set-optimised decision threshold used on that split. Test-fold sizes are 128, 127, 126, 128, 128 kernels for seeds 42, 123, 456, 789, 2024, of which the healthy class comprises 40, 23, 27, 26, 20 kernels respectively.

| **Split seed** | **Threshold** | **Balacc (%)** | **Healthy recall** | **Defect recall** | **False positives** | **False negatives** |
|---|---|---|---|---|---|---|
| 42 | 0.78 | 87.39 | 97.5% | 77.3% | 1 | 20 |
| 123 | 0.50 | 84.82 | 87.0% | 82.7% | 3 | 18 |
| 456 | 0.42 | 83.84 | 88.9% | 78.8% | 3 | 21 |
| 789 | 0.62 | 86.35 | 92.3% | 80.4% | 2 | 20 |
| 2024 | 0.52 | 88.89 | 100.0% | 77.8% | 0 | 24 |
| **mean ± std** | — | **86.26 ± 1.80** | **93.1% ± 5.0%** | **79.4% ± 2.0%** | **1.8 ± 1.2** | **20.6 ± 2.0** |